\documentclass[sigconf]{acmart}

\AtBeginDocument{%
  \providecommand\BibTeX{{%
    \normalfont B\kern-0.5em{\scshape i\kern-0.25em b}\kern-0.8em\TeX}}}

\usepackage{xspace}
\usepackage[utf8]{inputenc}
\usepackage{multirow}
\usepackage{multicol}
\usepackage{booktabs} % For formal tables
\usepackage{bm}
\usepackage{graphicx}
\usepackage{ftnxtra}
\usepackage{fnpos}
\usepackage{subcaption}
\usepackage[flushleft]{threeparttable}
\usepackage{flushend}
\usepackage{algorithm}
\usepackage{algpseudocode}
\usepackage{amsmath}
\usepackage{amsthm}
\usepackage{graphics}
\usepackage{bbm}

\usepackage{hyperref}
\copyrightyear{2022} 
\acmYear{2022} 
\setcopyright{acmcopyright}
\acmConference[CIKM '22]{Proceedings of the 31st ACM International Conference on Information and Knowledge Management}{October 17--21, 2022}{Atlanta, GA, USA}
\acmBooktitle{Proceedings of the 31st ACM International Conference on Information and Knowledge Management (CIKM '22), October 17--21, 2022, Atlanta, GA, USA}
\acmPrice{15.00}
\acmDOI{10.1145/3511808.3557097}
\acmISBN{978-1-4503-9236-5/22/10}

\newcommand{\DuMP}{DuMapper\xspace}
\newcommand{\DuMPOnline}{June 2018\xspace}
\newcommand{\ONEG}{\uppercase\expandafter{\romannumeral1}\xspace}
\newcommand{\SECG}{\uppercase\expandafter{\romannumeral2}\xspace}
\newcommand{\ONEGStar}{\uppercase\expandafter{\romannumeral1*}\xspace}
\newcommand{\SECGStar}{\uppercase\expandafter{\romannumeral2*}\xspace}
\begin{document}
\title{DuMapper: Towards Automatic Verification of Large-Scale POIs with Street Views at Baidu Maps}

\author{Miao Fan}
\email{fanmiao@baidu.com}
\orcid{0000-0003-1268-2937}
\affiliation{%
  \institution{Baidu Inc.}
  \streetaddress{Shangdi 10th Street}
  \city{Haidian District}
  \state{Beijing}
  \country{China}}

\author{Jizhou Huang}
\authornote{Corresponding author: Jizhou Huang.}
\email{huangjizhou01@baidu.com}
\orcid{0000-0003-1022-0309}

\affiliation{%
  \institution{Baidu Inc.}
  \streetaddress{Shangdi 10th Street}
  \city{Haidian District}
  \state{Beijing}
  \country{China}}

\author{Haifeng Wang}
\email{wanghaifeng@baidu.com}
\orcid{0000-0002-0672-7468}

\affiliation{%
  \institution{Baidu Inc.}
  \streetaddress{Shangdi 10th Street}
  \city{Haidian District}
  \state{Beijing}
  \country{China}}

\renewcommand{\shortauthors}{Miao Fan, Jizhou Huang, \& Haifeng Wang}

\begin{abstract}
With the increased popularity of mobile devices, Web mapping services have become an indispensable tool in our daily lives. To provide user-satisfied services, such as location searches, the point of interest (POI) database is the fundamental infrastructure, as it archives multimodal information on billions of geographic locations closely related to people's lives, such as a shop or a bank. Therefore, verifying the correctness of a large-scale POI database is vital. To achieve this goal, many industrial companies adopt volunteered geographic information (VGI) platforms that enable thousands of crowdworkers and expert mappers to verify POIs seamlessly; but to do so, they have to spend millions of dollars every year. 
To save the tremendous labor costs, we devised \DuMP, an automatic system for large-scale POI verification with the multimodal street-view data at Baidu Maps. This paper presents not only \DuMP \ONEG, which imitates the process of POI verification conducted by expert mappers, but also proposes \DuMP \SECG, a highly efficient framework to accelerate POI verification by means of deep multimodal embedding and approximate nearest neighbor (ANN) search. \DuMP \SECG takes the signboard image and the coordinates of a real-world place as input to generate a low-dimensional vector, which can be leveraged by ANN algorithms to conduct a more accurate search through billions of archived POIs in the database for verification within milliseconds. Compared with \DuMP \ONEG, experimental results demonstrate that \DuMP \SECG  can significantly increase the throughput of POI verification by $50$ times. \DuMP has already been deployed in production since \DuMPOnline, which dramatically improves the productivity and efficiency of POI verification at Baidu Maps. As of December 31, 2021, it has enacted over $405$ million iterations of POI verification within a 3.5-year period, representing an approximate workload of $800$ high-performance expert mappers.
\end{abstract}

\begin{CCSXML}
<ccs2012>
   <concept>
       <concept_id>10002951.10003227.10003236.10003237</concept_id>
       <concept_desc>Information systems~Geographic information systems</concept_desc>
       <concept_significance>500</concept_significance>
       </concept>
 </ccs2012>
\end{CCSXML}

\ccsdesc[500]{Information systems~Geographic information systems}

\keywords{web mapping services, POI verification, street views, deep multimodal embedding, approximate nearest neighbor search}

\maketitle

\section{Introduction}
\begin{figure}
\setlength{\abovecaptionskip}{0.2cm} 
    \centering
    \includegraphics[width=\columnwidth]{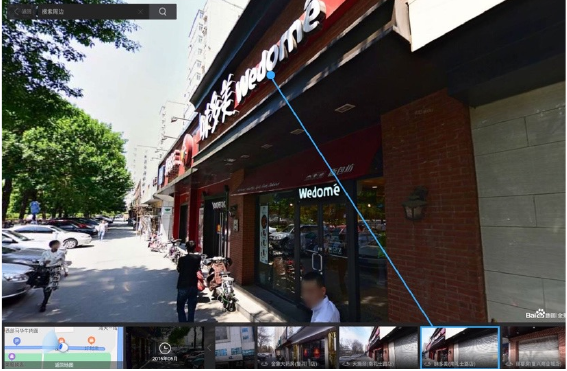}    \caption{A screenshot of the result of a POI search at Baidu Maps, a well-known Web mapping service in China. In this case, the user-desired POI has been verified by the street-view data where the signboard image, the name, and the coordinates of that POI are included.}
    \label{fig:poi_veri}
% \vspace{-2mm} 
\end{figure}

Ubiquitous mobile devices have greatly stimulated people's demand for location-based services \citep{10.1145/1325555.1325568,10.1145/953460.953490}, which consist of the capability to provide services to a user based on information about a given location. It makes Web mapping services \citep{10.1145/3394486.3403318,10.1145/3447548.3467058,10.1145/3394486.3412856,HGAMN2021,MetaPRec2021} become indispensable tools in our daily lives as more than $90\%$ of mobile owners have utilized them at least once to retrieve places they were interested in \citep{10.1145/1357054.1357125}. 
In order to provide user-satisfied search results in Web mapping services (see Figure~\ref{fig:poi_veri}),  point of interest (POI) databases \citep{chuang2016enabling,kim2014construction} are the fundamental infrastructures mainly because they can archive the multimodal information, such as the signboard (image), the name (text), and coordinates (numbers) on billions of real-world places, such as shops or banks, which are closely related to people's lives.
Therefore, it plays a vital role in Web mapping services to keep verifying the correctness of a large-scale POI database \citep{10.1145/2740908.2741715,ijgi10110735}. 

\begin{figure*}
\setlength{\abovecaptionskip}{0.2cm} 
\begin{subfigure}{\textwidth}
  \setlength{\abovecaptionskip}{0.15cm}
  \centering
  \includegraphics[trim=1.5cm 2.5cm 1.5cm 1.9cm, clip=true,width=0.9\textwidth]{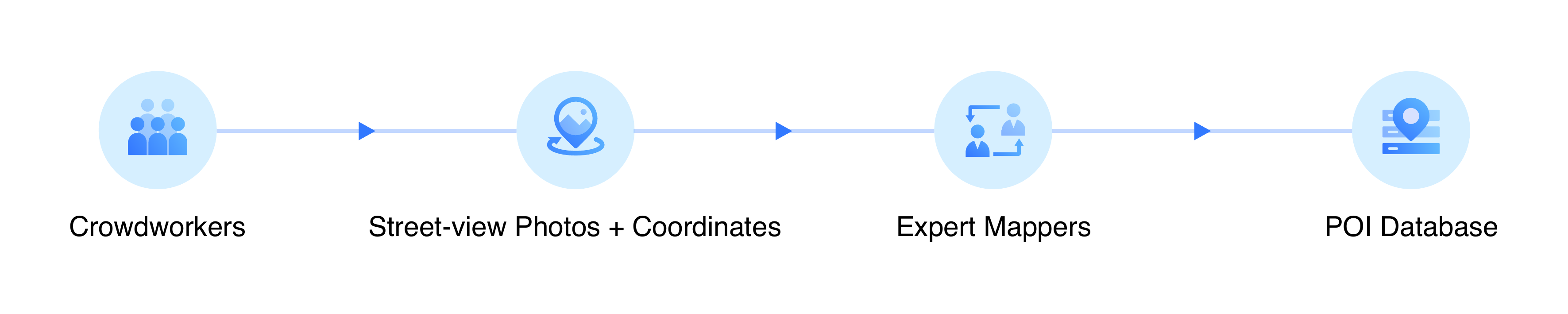}  
  \caption{The workflow of POI verification at the VGI platform at Baidu Maps.}
  \label{fig:vgi}
\end{subfigure}
\newline
\begin{subfigure}{\textwidth}
  \setlength{\abovecaptionskip}{0.15cm}
  \centering
  \includegraphics[trim=1.5cm 2.5cm 1.5cm 1.9cm, clip=true,width=0.9\textwidth]{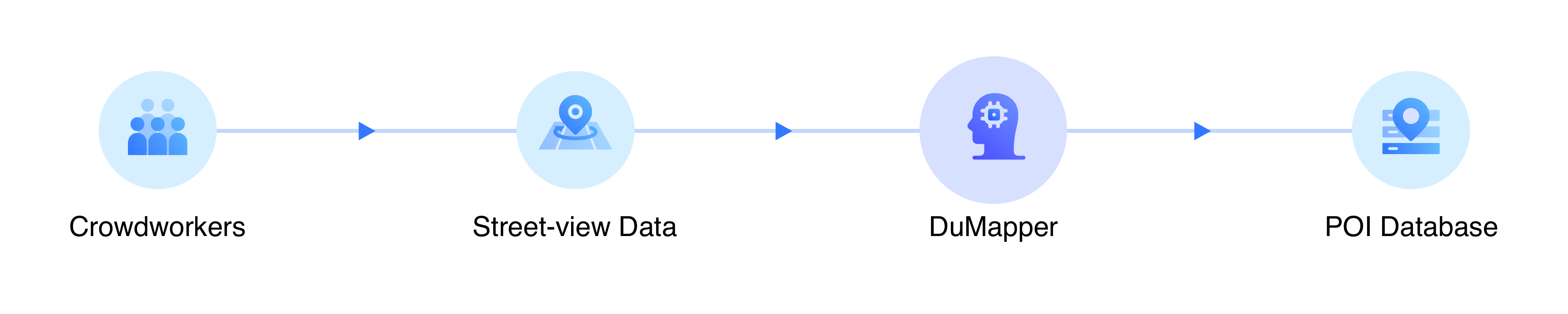}
  \caption{The workflow of automatic POI verification with the aid of \DuMP at Baidu Maps.}
  \label{fig:dumapper}
\end{subfigure}
  \caption{\DuMP is designed to be an intelligent agent that can continuously verify POIs with the multimodal street-view data at the VGI platform, in place of thousands of expert mappers.}
% \vspace{-2mm} 
\end{figure*}

However, it is challenging to keep a POI database in sync with the real-world counterparts, due to the dynamic nature of business changes and innovations. To ensure both the quantity and quality \citep{touya2017assessing} of the POI database, most industrial companies adopt volunteered geographic information (VGI) platforms \citep{goodchild2012assuring,coleman2009volunteered,flanagin2008credibility,jiang2015volunteered}, as illustrated by Figure \ref{fig:vgi}, which enable thousands of crowdworkers and expert mappers to verify POIs seamlessly. To be specific, crowdworkers are responsible for spotting POIs and submitting their street-view photos \citep{10.1145/3403931,ga2013new, BILJECKI2021104217} with coordinates to VGI platforms. Subsequently, expert mappers are in charge of retrieving a group of archived POIs close to each pair of crowdworker-submitted coordinates from the POI database. Then, they choose the same POI that appears in the street-view photo based on their knowledge and mapping expertise. To verify the correctness of a large-scale POI database, VGI platforms require tremendous labor costs. As both crowdworkers and expert mappers should be paid monthly, Web mapping services, such as Baidu Maps, had to spend millions of dollars every year in the past on the task of POI verification.  

Inspired by the recent breakthroughs on computer vision \citep{voulodimos2018deep,feng2019computer} and information retrieval with deep neural networks \citep{mitra2018introduction,10.1145/75335.75338}, we devised \DuMP, an automatic system for large-scale POI verification with the street views at Baidu Maps. As illustrated by Figure~\ref{fig:dumapper}, it is designed to be an intelligent agent that can continuously verify POIs with the multimodal street-view data at the VGI platform in place of thousands of expert mappers. Since \DuMPOnline, \DuMP has been deployed in production at Baidu Maps and dramatically improves the productivity of POI verification. As of December 31, 2021, it has enacted over 405 million iterations of POI verification within a 3.5-year period, which is equivalent to an approximate workload of 800 high-performance expert mappers. 

This paper first presents \DuMP \ONEG, the original automatic framework that imitates the process of POI verification conducted by expert mappers. To be specific, our framework decouples the street-view data into two parts: the coordinates where the street-view photo was taken, as well as the signboard images detected from this photo; and then it adopts a three-stage pipeline to deal with the coordinates and signboards for automatic POI verification, as follows.    

\begin{itemize}
    \item \textit{To fetch a group of POI candidates through geo-spatial index} \citep{van2015efficient}: This module takes the coordinates where the street-view photo was taken as input and retrieves a group of POI candidates located within a radius of $r$ kilometers with the coordinates as the center. The radius $r$ is a heuristic argument that is leveraged to fix the deviation between the shooting position of the photo and the actual coordinates of the target POI. At the same time, the number of candidate POIs has also significantly been reduced from billions to thousands.

    \item \textit{To obtain the name of the target POI utilizing optical character recognition} \citep{Chaudhuri2017}: In this module, OCR technology is used to recognize the name from the signboard image of the target POI in the street-view photo.
    
    \item \textit{To find the target POI via ranking candidate POIs with the evidence of multimodal features} \citep{gomaa2013survey}: This module conducts the calculation of multimodal similarity between the collected POI and each candidate POI one-by-one, and the candidate POI with the highest similarity is selected from the POI database for verification. 
\end{itemize}

Although \DuMP \ONEG has automated the workflow of expert mappers for POI verification with the aid of geo-spatial index (GSI), optical character recognition (OCR), and candidate POI rank (CPR), the efficiency bottleneck of this framework still exists. Given that the group of candidate POIs fetched by GSI is dynamically changed along with the coordinates submitted by crowdworkers, it leads to an issue that the thousands of candidate POIs cannot be pre-indexed, and to find the target POI among them is hardly accelerated. In other words, this problem is caused by the separation of indexing coordinates and ranking candidate POIs. 

To address this problem, we propose a novel framework, \DuMP \SECG, which can directly index the multimodal street-view data (including the signboard image and the coordinates of a POI) and highly accelerate the speed of automatic POI verification. It is mainly composed of two advanced modules, as follows.

\begin{itemize}
    \item \textit{Deep multimodal embedding (DME)} \citep{CHEN2021195} \textit{of a POI}: This neural module takes the signboard image and the coordinates of a POI as input to generate its multimodal vector representation. In this way, both the archived POIs in the database and a real-world place captured by the street-view data can be mapped into the same low-dimensional space. 
    \item \textit{Approximate nearest neighbor (ANN) search}  \citep{10.5555/2976040.2976144} \textit{through the large-scale POI database}: This module takes advantage of ANN algorithms to conduct a more accurate search through billions of archived POI embeddings in the database for verification within milliseconds.
\end{itemize}

This novel framework revolutionizes the means of large-scale POI verification and greatly improves the efficiency of \DuMP \ONEG. Compared with \DuMP \ONEG, experimental results demonstrate that \DuMP \SECG can significantly increase the throughput of automatic POI verification at Baidu Maps by $50$ times. 
For reproducibility tests, we have released the source code of \DuMP \SECG.\footnote{The code is available at \url{https://github.com/PaddlePaddle/Research/tree/master/ST_DM/CIKM2022-DuMapper/}.}

\section{Related Work}
In this section, we give an overview of the related work on POI verification in Web mapping services. Given that POI verification is the key task of POI database construction, and that constructing a POI database canonically relies on VGI systems, we will focus here on the two research themes: volunteered geographical information and POI database construction.

\subsection{Volunteered Geographical Information}
Volunteered geographical information (VGI) systems encourage the aid of volunteers to generate maps. It involves crowdsourcing workers in producing a highly detailed world map which was only followed by mapping agencies and other government authorities. A representative example of VGI is OpenStreetMap \citep{4653466}, an open and collaborative platform where anyone can contribute and edit POIs. Several studies \citep{doi:10.1080/13658816.2013.867495,4653466} state that OpenStreetMap can lead to high completeness and correctness but a fair position accuracy. Although OpenStreetMap follows the collaborative paradigm on content generation, which Wikipedia also adopts, the main limitation of the platform is exposed as people cannot contribute remotely to a specific POI.

To address the problem, Virtual City Explorer (VCE) \citep{10.1145/3403931} has been recently proposed, wherein participants are not required to be physically present in a specific POI. It enables expert mappers to collect coordinates of POIs using digital street-view imagery. Inspired by VCE, we found out that the workload of thousands of expert mappers for POI verification at Baidu Maps may be further saved by an intelligent agent, where billions of archived POIs can be seamlessly verified with the multimodal street-view data. 

\subsection{POI Database Construction}
Over the past decades, OpenStreetMap \citep{4653466} and Wikimapia \citep{doi:10.1080/13658816.2018.1463441} have been the representative POI databases mainly established by VGI systems, wherein hundreds of thousands of contributors have collected millions of points of interest, including towns, restaurants, lakes, and tourist attractions. Given that crowdsourcing is their major force for POI generation, the confidence in POI data quality from user-generated content has always been an essential issue on POI database construction by means of VGI systems.

To increase the credibility of information sources for POI generation, several studies \citep{10.1145/2740908.2741715,chuang2016enabling,10.1145/3459637.3481924} have been focused on location entity extraction from unstructured Web pages, such as Internet news and official
websites. It is because many location entities prefer to publish the business information on their official websites or Internet news in a timely fashion, demonstrating that they are valuable data sources for large-scale extraction of POI information. With the help of Web crawling \citep{10.14778/1454159.1454163} and information extraction \citep{10.1145/1409360.1409378,DuIVRS2022} techniques, a POI database can be automatically constructed. Although obtaining POI data from the Internet solves the problem of information credibility to a certain extent, data coverage can become another problem.

Compared with the text information of POIs on the Web, the signboards appearing in the street-view photos are more widely distributed across the world. Motivated by the broad availability of geo-tagged street-view photos, recent studies \citep{ga2013new,8954142} proposed a new task on map maintenance via automatically detecting POIs from street views. The new task generally makes the update process of POI databases more proactive because it gives us the same view as we would see on the street in the real world. To be specific, the street-level imagery has a distinctive strength in POI database construction, as not only does the signboard image in the street-view photo imply the existence of a POI, but also the name, location, and address extracted from the street-view data can be used to fill in the POI database. 

Although we initially devised \DuMP \ONEG, which partly adopts the approaches \citep{ga2013new,8954142} on POI verification with street views, it still lacks efficiency as the dynamic group of candidate POIs are hardly pre-indexed. Therefore, we further contribute \DuMP \SECG to both research and industrial communities, which revolutionizes the method of obtaining large-scale POI verification by neural information retrieval \citep{mitra2018introduction,GUO2020102067} and greatly improves the efficiency of automatic POI verification with the multimodal street-view data.

\section{D\texorpdfstring{\MakeLowercase {u}}{u}M\texorpdfstring{\MakeLowercase{apper}}{apper} {\romannumeral1}}
\begin{figure}
\setlength{\abovecaptionskip}{0.2cm} 
    \centering
    \includegraphics[trim=0.0cm 0.6cm 2.0cm 2.1cm, clip=true,width=0.85\columnwidth]{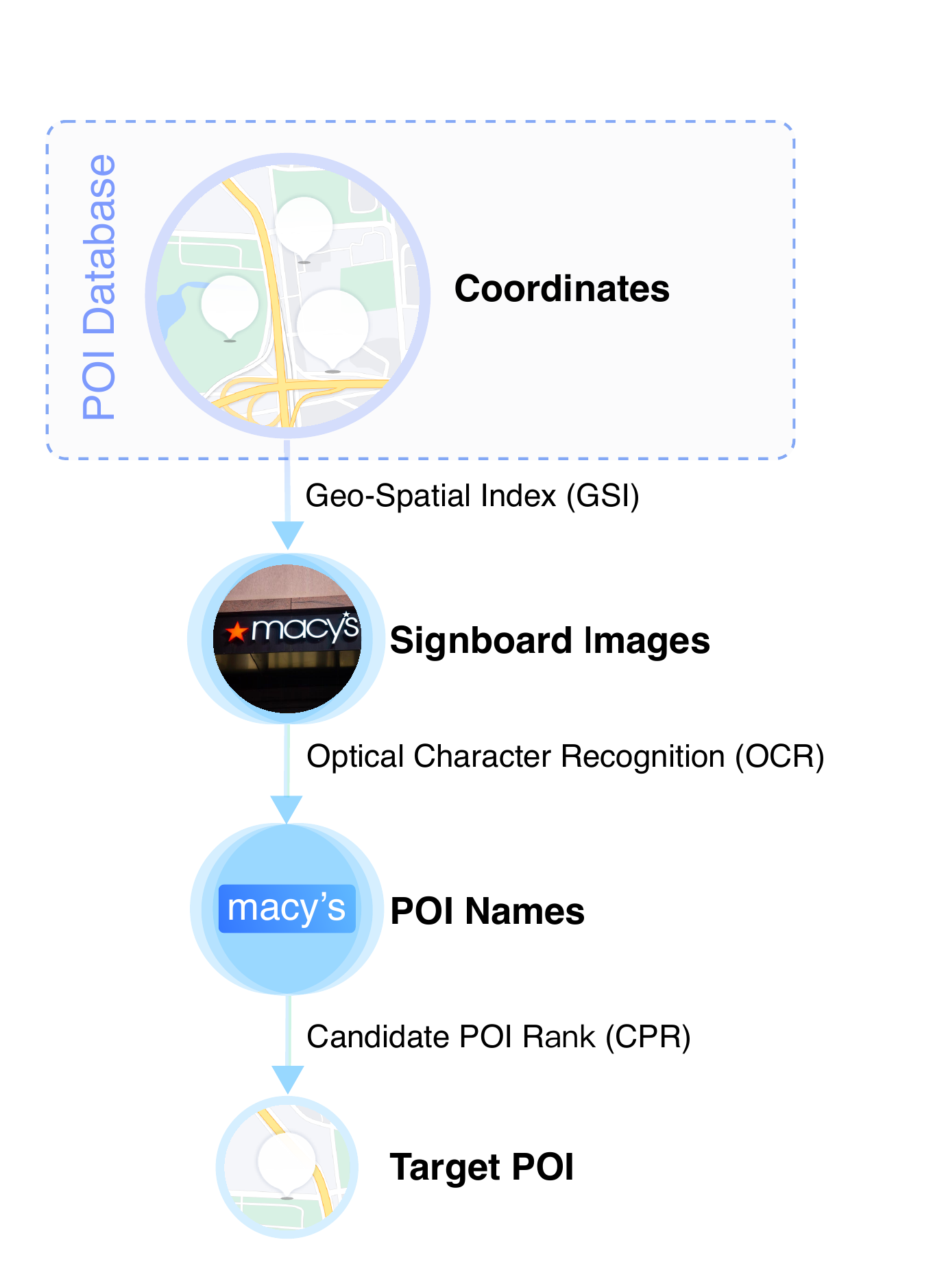}
    \caption{\DuMP \ONEG is the original automatic framework which imitates the process of POI verification conducted by expert mappers at Baidu Maps. It adopts a three-stage pipeline, i.e., geo-spatial index (GSI), optical character recognition (OCR), and candidate POI rank (CPR), to deal with the coordinates, as well as the signboards in street-view data for automatic POI verification.}
    \label{fig:dumap1}
% \vspace{-2mm}
\end{figure}

\DuMP \ONEG  imitates the work method conducted by expert mappers for automatic POI verification at Baidu Maps. As illustrated by Figure~\ref{fig:dumap1}, it decouples the street-view data into two parts: the coordinates where the street-view photo was taken, as well as the signboards in this photo and then it adopts a three-stage pipeline (i.e., geo-spatial index, optical character recognition, and candidate POI rank) to successively deal with the coordinates and signboard images for automatic POI verification.  

\subsection{Geo-Spatial Index}
\subsubsection{Problem Statement}
The key challenge of \DuMP \ONEG  is how to accurately find the only one POI which appears in the street-view photo from billions of archived POIs at Baidu Maps. On the one hand, the time complexity is extremely high to compare a signboard image one-by-one with the photos of large-scale POIs in the database. On the other hand, the coordinates just show the exact location where the crowdworkers submitted street-view photos but cannot be used directly to find the location of the target POI, as there is also a distance between these two locations which is difficult to measure precisely. Therefore, it becomes the first problem of \DuMP \ONEG  to choose to which kind of data to narrow the scope of POI retrieval.

\subsubsection{Practical Solution}
To address the problem, we decide to use the coordinates as the geo-spatial index to narrow the scope of POI retrieval. The main reasons for this practical solution are two-fold. On the one hand, the coordinate features are more lightweight compared with the features of the signboard images, and the space and time consumption of the index are less.
On the other hand, POI images can provide more accurate matching features than coordinates and are more suitable for ranking modules. We leverage ArangoDB to support geo-spatial index, which supersedes other geo-index implementations with the aid of Google S2 geocoding method \citep{ekawati2018analysis}. This index assumes coordinates with the latitude between $-90$ and $90$ degrees, as well as the longitude between $-180$ and $180$ degrees, and it will ignore all the archived POIs whose spatial distance from the query coordinates is greater than $r$ kilometres.

\subsection{Optical Character Recognition}
\subsubsection{Problem Statement}
After we have obtained thousands of candidate POIs via narrowing the search scope by the geo-spatial index, how to select only one accurate POI from these candidate POIs becomes one of the key problems of ranking. Although the signboard contains rich image and text information compared with coordinates, it is a difficult problem to extract key features for POI ranking, due to the limitations of shooting angle, illumination, etc. 
These extracted key features can be the name of POI or even the outline information of some signboard images. 
Therefore, whether to choose text or image features as the evidence for POI ranking and how to extract the corresponding features have become the key issues at this stage.

\subsubsection{Practical Solution}
We consider text information to be less susceptible to the shooting angle and illumination change. After we decide to adopt text as the key evidence for ranking, how to extract the textual names of POIs from the signboard images becomes the primary problem. To address the issue, optical character recognition (OCR) is a well-studied technique wherein powerful algorithms already exist. However, the street-view photos taken by portable camera devices do not only contain the signboards of POIs. Therefore, we adopt the conventional steps of text extraction from natural scene images \citep{zhang2013text} as the practical solution to key evidence generation before POI ranking, which is generally composed of three stages: (1) signboard detection and localization; (2) text enhancement and segmentation; and (3) optical character recognition.
 
\subsection{Candidate POI Rank}
\subsubsection{Problem Statement}
Given that the results of OCR have a certain probability of being wrong, the problems faced by this module are also different. On the one hand, we need to figure out the way to correct the POI name that was incorrectly recognized from a signboard image. On the other hand, multiple POIs with the same name may also be selected from the set of candidate POIs retrieved by GSI, even if the POI name is correctly extracted by OCR from a signboard image. Therefore, how to further deal with the duplicate name problem is also one of the issues of this stage.

\subsubsection{Practical Solution}
The key to reducing the mistakenly recognized samples of OCR is to find the patterns whereby it goes wrong. However, an off-the-shelf OCR software is usually a black box to us: We can only observe its inputs and outputs. Therefore, we decide to train an error correction model to refine the POI names generated by OCR. In order to obtain a large amount of parallel corpus for training, we use the off-the-shelf OCR software to scan all the archived POIs which contain signboard images. As we also know the correct names of those POIs, a seq2seq model \citep{ge2019automatic} can be established for the task of error correction of POI names, with the capability of modelling the frequent patterns on the defects of OCR, based on massively parallel corpora. In this way, the refined name of the recognized POI from signboard images can be used to compute the similarities with the names of candidate POIs. For the problem of duplicated names in the set of candidate POIs, we introduce the outline features of signboard images into the ranking model to distinguish POIs just when the ranked POIs at the top contain the same name, so as to significantly reduce unnecessary feature computations.

\section{D\texorpdfstring{\MakeLowercase {u}}{u}M\texorpdfstring{\MakeLowercase{apper}}{apper} {\romannumeral2}}
Although \DuMP \ONEG has automated the work of expert mappers at VGI platforms, the efficiency bottleneck of this original framework still exists. The main reason is that the group of candidate POIs obtained from the geo-spatial index is dynamically changed, along with the coordinates submitted by crowdworkers. It leads to an issue: that the thousands of candidate POIs cannot be pre-indexed. As a result, finding the target POI among them is hardly accelerated.  

\begin{figure}
\setlength{\abovecaptionskip}{0.2cm} 
    \centering
    \includegraphics[trim=1.0cm 1.0cm 1.0cm 1.0cm, clip=true,width=\columnwidth]{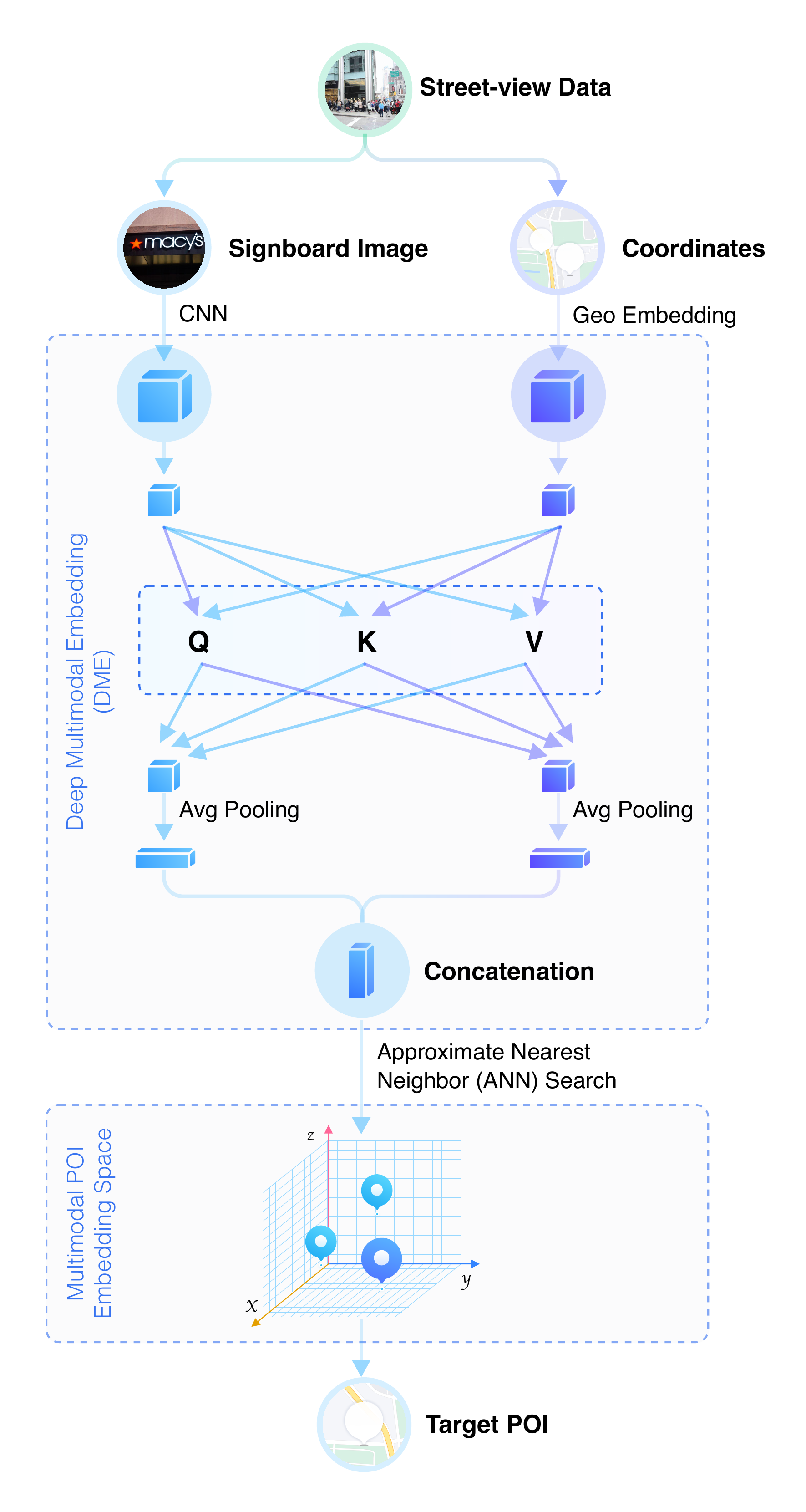}
    \caption{\DuMP \SECG revolutionizes the way to attain large-scale POI verification by means of two advanced modules: deep multimodal embedding (DME) of a POI and approximate nearest neighbor (ANN) search through the large-scale POI database. This novel framework can highly accelerate the speed of automatic POI verification.}
    \label{fig:dumap2}
% \vspace{-5mm}
\end{figure}

To address this issue, we propose a novel framework \DuMP \SECG, illustrated by Figure~\ref{fig:dumap2}, which is composed of two advanced modules: deep multimodal embedding (DME) and approximate nearest neighbor (ANN) search. It revolutionizes the way to attain large-scale POI verification by neural information retrieval \citep{mitra2018introduction} and greatly improves the efficiency of automatic POI verification.

\subsection{Deep Multimodal Embedding}
\subsubsection{Problem Statement}
The major challenge of~ \DuMP \ONEG, which limits the efficiency of automatic POI verification, is the serial computing mode of three different modal features, including the coordinates, the signboard image, and the text name of a POI. Therefore, the first and most critical step of \DuMP \SECG  is to fuse the features of the above three different modalities at the same time. In this way, each archived POI in the database can be represented by a multimodal feature and pre-indexed by some kind of data structure (such as trees or graphs) for efficient search. 
Even if the crowdworkers daily submit millions of requests on POI verification, \DuMP \SECG  can easily handle this huge volume of requests. In order to achieve this goal, we need to design an encoding mechanism for the multimodal features of a POI.

\subsubsection{Practical Solution} 
The street-view data submitted by crowdworkers mainly contain the signboard images of POIs along with their coordinates. A pair of coordinates is conventionally a two-dimensional vector. It contains the longitude ($x$) and the latitude ($y$), where the numerical range of longitude is $(-180.000000, 180.000000)$ and that of latitude is $(-90.000000, 90.000000)$. However, a signboard image is usually stored by a matrix or even a tensor. Thus, the cross-modal data with two different dimensions cannot be directly merged.
To address this issue, we build a neural model that takes the signboard image, as well as the coordinates of a POI, as input and learns to generate its multimodal vector representation. In this way, not only the archived POIs in the database, but also the real-world places captured by crowdworkers, can be mapped into the same low-dimensional space. 

As one of the inputs of the neural model, we use a canonical CNN (convolutional neural net) \citep{lecun1995convolutional} to extract the feature matrix ($\textbf{G}'$) from the raw data (denoted by $\textbf{Z}$) of the signboard image, as follows: 
\begin{equation}
    {\textbf G}' = \text{CNN}(\textbf{Z}),
\end{equation}
where $\textbf{G}' \in \mathbb{R} ^ {l \times d}$.

For the other source of input, i.e., the coordinates $(x, y)$, we first use the Geohash algorithm \citep{moussalli2015fast} to map the coordinates to a discrete vector space (denoted by ${\textbf i'} \in \mathbb{R} ^ {l}$):
\begin{equation}
    {\textbf i'} = \text{GeoHash}(x, y),
\end{equation}
and we further expand each dimension of ${\textbf i'}$ to a random embedding:
\begin{equation}
    {\textbf I'} = \text{GeoEmb}(\textbf{i}'),
\end{equation}
where $\textbf{I}' \in \mathbb{R} ^ {l \times d}$.

Then we devise a cross-attention neural network to fuse the two kinds of features:
\begin{equation}
    {\textbf I} = \text{Softmax}(\frac{\textbf{Q}_{G}\textbf{K}_{I}^T}{\sqrt{d}})\textbf{V}_{I},
\end{equation}
where $\textbf{Q}_{G} = \textbf{W}^T\textbf{G}'$, $\textbf{K}_I = \textbf{V}_I = \textbf{U}^T\textbf{I}'$,
and
\begin{equation}
    {\textbf G} = \text{Softmax}(\frac{\textbf{Q}_{I}\textbf{K}_{G}^T}{\sqrt{d}})\textbf{V}_{G},
\end{equation}
where $\textbf{Q}_{I} = \textbf{U}^T\textbf{I}'$, $\textbf{K}_G = \textbf{V}_G = \textbf{W}^T\textbf{G}'$.

Finally, we conduct average pooling on ${\textbf I}$ and ${\textbf G}$, respectively:
\begin{equation}
    {\textbf i} = \text{Avg\_Pool}(\textbf{I}),
\end{equation}
and
\begin{equation}
    {\textbf g} = \text{Avg\_Pool}(\textbf{G}).
\end{equation}

As a result, the deep multimodal embedding of a POI can be represented by:
\begin{equation}
    {\textbf m} = {\textbf i} \oplus {\textbf g},
\end{equation}
where $\oplus$ is the operator of concatenation for two vectors.

In order to learn the parameters (such as $\textbf{W}$, $\textbf{U}$, etc.) of this model, we need a training set $\Delta$ where each POI has multiple street-view data submitted by crowdworkers. The objective is to minimize the sum of the triplet loss $\mathcal{L}_{\Delta}$ over the training set, which is defined as follows:
\begin{equation}
    \mathcal{L}_{\Delta} = \sum_{(m, m_+, m_*) \in \Delta}{\max \{0, \gamma + \frac{\textbf{m}~\boldsymbol{\cdot}~ \textbf{m}_{+}}{|\textbf{m}| |\textbf{m}_{+}|} - \frac{\textbf{m}~\boldsymbol{\cdot}~ \textbf{m}_{*}}{|\textbf{m}| |\textbf{m}_{*}|}}\},
\end{equation}
where $\textbf{m}$ and $\textbf{m}_+$ belong to the same POI, but $\textbf{m}_*$ is randomly sampled from other POIs. Moreover, $\gamma$ is the hyperparameter standing for the margin.

\subsection{Approximate Nearest Neighbor Search}
\subsubsection{Problem Statement}
The original task of this step is to find the target POI through the archived POIs in the database queried by the POI in the street-view photo. However, after we have obtained the neural model for the multimodal embedding of POIs, both the POI in the street-view photo and the archived POIs can be represented by low-dimensional vectors, which are projected into the same geometric space. Thus, the problem of this step can be stated as a formal expression that, given a set of POI embeddings from the database (denoted by $P$), we need to build a data structure that returns the nearest neighbor $p \in P$ of any query POI (denoted by $q$) in the street-view data.

\subsubsection{Practical Solution}
A simple solution to this problem would store the set $P$ in memory, compute all distances $D(q, p)$ for $p \in P$, and select the point $p$ with the minimum distance. Obviously, it is impractical to calculate all the $n$ distances, as they require at least $n$ operations. Since, in POI databases, $n$ can be as large as $10^9$, it was necessary to develop faster methods that can find the nearest POI without explicitly computing all distances from the query POI in the street-view photo. To address the issue, we use Milvus\footnote{\url{https://milvus.io/}} as the vector database to build an efficient index for scalable similarity search. Moreover, ANNOY\footnote{\url{https://github.com/spotify/annoy}} is adopted as the index approach because it can divide the embedding space into multiple subspaces and store the vectors of archived POIs in a tree structure. To be specific, ANNOY allows the query vector to follow the tree structure to keep finding the subspace which is closest to the target vector. Then, we just need to compare the distances between the query vector and all the vectors in the subspace, which is on a leaf node of the tree, to obtain the target POI. As a result, the lookup complexity can be reduced logarithmically. 

\section{Experiments}
As an intelligent agent for industrial use of automatic POI verification, \DuMP needs to be tested thoroughly, regardless of offline or online status. In this section, we will report the results of extensive experiments wherein the performance of several frameworks on POI verification was evaluated by multiple metrics.

\subsection{Comparison Frameworks}
Besides the VGI platform supported by expert mappers, there are two versions of \DuMP for POI verification at Baidu Maps. For each version, we have optimized and upgraded it appropriately. Therefore, a total of five different frameworks were involved in the experimental evaluation.

\subsubsection{Expert Mapper (VGI)}
In order to obtain a benchmark assessment on both accuracy and efficiency of POI verification conducted by a professional worker, we monitored the production capacity of more than $100$ expert mappers employed by Baidu Maps over $2$ weeks and report the average performance of one expert. As expert mappers do not need to participate in the offline evaluation, the results of the assessment will serve as comparison data in the online A/B test for automatic POI verification.

\subsubsection{\DuMP \ONEG} This is the first automatic framework adopted by Baidu Maps for automatic POI verification. It imitates the workflow conducted by expert mappers and adopts a three-stage pipeline, i.e., geo-spatial index (GSI), optical character recognition (OCR), and candidate POI rank (CPR), to deal with the coordinates, as well as the signboard images, in street-view data, for the task of automatic POI verification.

\subsubsection{\DuMP \ONEGStar} Inspired by the recent work on automatic POI verification with street views \citep{ga2013new,8954142}, \DuMP \ONEGStar further leverages the matching feature of signboard images, which may facilitate POI search by the additional information. Superior to \DuMP \ONEG, it upgrades the CPR module via extending the modality of ranking features and is regarded as the state-of-the-art framework on automatic POI verification.

\subsubsection{\DuMP \SECG} This is a novel framework contributed by this paper for automatic POI verification. It revolutionizes the method of \DuMP \ONEG which only imitates the workflow conducted by expert mappers. To increase the productivity of POI verification, \DuMP \SECG is composed of two advanced modules: projecting all the archived POIs into the same vector space by deep multimodal embedding (DME), and directly obtaining the nearest POI for verification via an approximate nearest neighbor (ANN) search. 

\subsubsection{\DuMP \SECGStar} Although \DuMP \SECG  is extremely efficient with the aid of ANN search, it is prone to reduce the accuracy of POI retrieval if we just set the number of nearest neighbor to $1$. As an industrial framework for automatic POI verification, accuracy is the most important indicator. Therefore, we made a trade-off between accuracy and efficiency of \DuMP \SECG and further proposed \DuMP \SECGStar. The number of nearest neighbors is set to $10$ in the ANN search, and we bring the multimodal ranking module in \DuMP \ONEGStar as the final decision step for POI verification. 

\subsection{Offline Assessment}
\subsubsection{Real-world Dataset}
The VGI platform was the mainstream source to keep the POI database of Baidu Maps in sync with its real-world counterparts. Large-scale street-view data have been submitted by thousands of crowdworkers over years. For offline assessment, we randomly sampled a real-world dataset from the POI database, where each instance of data is mainly composed of the identifier of a POI and the signboard images of the POI, as well as the coordinates where the signboard images were taken.
Table~\ref{tab:offline_dataset} shows statistics of the real-world dataset for the offline assessment of automatic POI verification at Baidu Maps. There are three subsets in the dataset, which are separately leveraged for the purpose of model training (abbr. \textit{Train}), hyper-parameter tuning (abbr. \textit{Valid}), and performance testing (abbr. \textit{Test}).

We can also observe from Table~\ref{tab:offline_dataset} that most POIs have about $6$ signboard images, with different coordinates where the images were taken. Given that the POIs in the training and validation subsets are the same, the average number of signboard images with coordinates are approximately $4$ and $2$ in the training and validation subsets, respectively. To test the generalizability of different frameworks on automatic POI verification, the POIs in the test subset are completely different from those in the training and validation set. This explains why the average number of signboard images with coordinates is also about $6$ in the test subset. 

\begin{table}[!t]
    \begin{center}
        \caption{The statistics of the real-world dataset for the offline assessment of automatic POI verification. The subsets are separately leveraged for the purpose of model training (abbr. {\it Train}), hyper-parameter tuning (abbr. {\it Valid}), and performance testing (abbr. {\it Test}). Each instance of data is mainly composed of the identifier of a POI and the signboard images of the POI, as well as the coordinates where the signboard images were taken.} 

        \begin{tabular}{l|rrr}
            \toprule
            \textbf{Subset} &  \textbf{\#(POIs)} &  \textbf{\#(Signboards)} & \textbf{ \#(Coordinates)} \\
            \hline
            \hline
          
            \textit{Train}  & $50,000$ &$200,050$ & $200,050$ \\
            \textit{Valid}  &$50,000$ & $96,900$ & $96,900$ \\
            \textit{Test}   &$12,000$ & $73,218$ & $73,218$ \\

            \bottomrule
            
        \end{tabular}
        
        \label{tab:offline_dataset}
    \end{center}
\end{table}

\subsubsection{Evaluation Metrics}
\label{sec:off_metric}
For each signboard image with coordinates, the task of POI verification is to find the exact POI through the large-scale POI database with the highest probability. It leads us to use SR@K as the metric to evaluate the success rate of ranking the target POI in top K predictions. Although SR@1 is the desired measurement which is also mainly concerned with online testing, we still decide to determine the results of the offline assessment on SR@3 and SR@5. The purpose of this is to know the proportion of target POI ranked from second to fifth, so as to guide our optimization direction of automatic POI verification.

\begin{table}
    \begin{center}
        \caption{The experimental results of the offline evaluations on different frameworks for automatic POI verification at Baidu Maps. All the frameworks are tested by the real-world dataset shown by Table~\ref{tab:offline_dataset} and measured by the success rate to rank the target POI in top K predictions (i.e., SR@K).} 
    
        \begin{tabular}{l|rrr}
            \toprule
            \multirow{2}{*}{\textbf{Framework}} & \multicolumn{3}{c}{\textbf{Evaluation Metrics (Offline)}}\\
            & {\textbf{SR@1}} & {\textbf{SR@3}} & {\textbf{SR@5}}   \\
            \hline
            \hline
      
            \textit{\DuMP \ONEG}  & $78.42\%$ & $80.15\%$ & $85.69\%$ \\
            \textit{\DuMP \ONEGStar}  & $87.66\%$ & $89.23\%$ & $93.38\%$ \\
\hline
            \textit{\DuMP \SECG}  & $88.08\%$ & $93.17\%$ & $95.12\%$ \\
            \textit{\DuMP \SECGStar}  & $91.74\%$ & $92.89\%$ & $95.46\%$ \\
            \bottomrule
        \end{tabular}
        
        \label{tab:offline_eva}
    \end{center}
\end{table}

\subsubsection{Analysis of Experimental Results}
Table~\ref{tab:offline_eva} reports the experimental results of the offline evaluations on different frameworks for POI verification at Baidu Maps. All the automatic frameworks are trained and validated, as well as tested by the real-world dataset shown by Table~\ref{tab:offline_dataset} and measured by SR@K. 
From the results of offline assessments, we can see that 2nd-generation DuMappers (i.e., \DuMP \SECG  and \DuMP \SECGStar) generally perform better than those of 1st-generation. Compared with \DuMP \ONEG, the state-of-art framework \DuMP \ONEGStar obtains significant improvements by $9.24\%$ SR@1, $9.08\%$ SR@3, and $7.69\%$ SR@5, indicating that the features of signboard images play a huge role in automatic POI verification. Though \DuMP \SECG outperforms \DuMP \ONEGStar, there exists a gap between the result on SR@1 and SR@3 in \DuMP \SECG. It indicates that over $5\%$ ground-truth POI ranked second and third. Therefore, the re-ranking module in \DuMP \SECGStar further improves the accuracy of \DuMP \SECG  and achieves the highest SR@1 of $91.74\%$.

\subsection{Online A/B Testing}
\subsubsection{Traffic of Street-view Data}
We have updated the workflow of POI verification at Baidu Maps multiple times. For each time, we would routinely deploy the new framework online and make it randomly serve 10\% traffic of the street-view data submitted by thousands of crowdworkers. During the period of A/B testing, the performance of the new framework would be monitored and compared with the performance of the framework that has already been launched to production. This period conventionally lasts for at least one week. Even if the results of A/B testing show that the new framework can outperform the one in production, we still need to keep an eye on the performance of the new framework when it begins to serve the full traffic of street-view data for some time, in case of the bias on traffic sampling. 

\subsubsection{Evaluation Metrics}
As an industrial framework for automatic POI verification, accuracy is the most important indicator which is equivalent to SR@1. After all, what is ultimately required by the users who daily use Web mapping services is the correctness of POI information. While trying to ensure the accuracy of POI information, we also hope to reduce costs and increase efficiency. Therefore, the other one of the critical objectives of the task of POI verification is to increase its throughput. To quantify this throughput, we borrow the QPS (query per second) indicator from information retrieval systems. The indicator implies the efficiency of an automatic POI verification framework, representing how many POI verification requests the framework can handle per second on average.

\begin{table}
    \begin{center}
        \caption{The experimental results of the online A/B testing on different frameworks for automatic POI verification at Baidu Maps. All the frameworks are tested by the real traffic of the street-view data submitted by the crowdworkers. Except for the measurement of SR@K for offline assessments, they can be further evaluated by another two indicators on both accuracy and efficiency of production.} 
        \begin{tabular}{l|rr}
            \toprule
            \multirow{2}{*}{\textbf{Framework}} & \multicolumn{2}{c}{\textbf{Evaluation Metrics (Online)}}\\ 
            & \textbf{Accuracy}
            & \textbf{QPS} \\
            \hline
            \hline

                \textit{\DuMP \ONEG}  &$75.16\%$ & $3.02$ \\
                \textit{\DuMP \ONEGStar}  &$84.79\%$ & $1.67$ \\
                \hline
                \textit{\DuMP \SECG}  &$85.14\%$ & $152.49$ \\
                \textit{\DuMP \SECGStar}  &$90.89\%$ & $7.38$ \\
                \hline
                \textit{Expert Mapper (VGI)}  &$94.52\%$ & $0.007$ \\

            \bottomrule
        \end{tabular}
        
        \label{tab:online_eva}
    \end{center}
% \vspace{-3mm}    
\end{table}

\subsubsection{Analysis of Experimental Results}
Table~\ref{tab:online_eva} shows the experimental results of the online A/B testing on different frameworks for automatic POI verification. All the frameworks are tested by the real traffic of the street-view data submitted by the crowdworkers at Baidu Maps. Although accuracy is the most important indicator, which is equivalent to SR@1, QPS is another key objective to reduce costs and increase efficiency. Therefore, the other one of the critical objectives of POI verification task is to increase its throughput. 
From the results of online testing, we can first figure out that an expert mapper performs the highest accuracy of $94.52\%$, without a doubt, but dramatically slower than automatic systems. \DuMP \SECG  owns the highest QPS of $152.49$, which is 50 times of the efficiency of \DuMP \ONEG. The throughput of the automatic POI verification framework tends to be significantly reduced by the re-ranking module which is widely adopted by \DuMP \ONEG, \ONEGStar, and \SECGStar; however, the module can help increase the accuracy. Therefore, we made a trade-off between accuracy and efficiency in \DuMP \SECGStar by reusing the multimodal ranking module in \DuMP \ONEGStar as the final decision step for POI verification and achieved the highest accuracy among all the automatic POI verification frameworks. Even so, the efficiency of \DuMP \SECGStar can still be at least doubled, since the number of POIs used for re-ranking is fixed and controllable.

\section{Conclusion}
In this paper, we present \DuMP, an automatic system for large-scale POI verification with the multimodal street-view data at Baidu Maps. Our critical contributions to both research and industrial communities are listed as follows.

\begin{itemize}
    \item \textit{Potential impact}: \DuMP is designed to be an intelligent agent that can continuously verify POIs with the multimodal street-view data at the VGI platform in place of thousands of expert mappers. It plays a vital role in Web mapping services to keep a large-scale POI database in sync with its real-world counterparts.

    \item \textit{Novelty}: \DuMP \SECG revolutionizes the original way of enacting automatic POI verification, i.e., \DuMP \ONEG, which just imitates the workflow conducted by expert mappers. \DuMP \SECG is a highly efficient framework, which is mainly composed of two advanced modules; (1) A deep multimodal embedding module (DME),  which can take the signboard image and the coordinates of a POI from the street-view data as input and generate a multimodal vector representation; and (2) An approximate nearest neighbor (ANN) search module, which takes advantages of ANN algorithms to conduct a more accurate search through billions of archived POI embeddings in the database for verification within milliseconds.
    
    \item \textit{Technical quality}: Extensive assessments have been conducted both offline and online. Compared with \DuMP \ONEG, experimental results demonstrate that \DuMP \SECG can significantly increase the throughput of automatic POI verification by $50$ times.
    
    \item \textit{Scalability}: The performance of \DuMP is equivalent to the workload of $800$ high-performance expert mappers.
    
    \item \textit{Reproducibility}: 
    We have released the source code of \DuMP \SECG  to ensure the reproducibility of this work.
\end{itemize}

\section{Future Work}
POI verification is not the only way to maintain a POI database's correctness. Statistics show that the attributes of $74.5\%$ POIs at Baidu Maps have been updated in the year 2020 \citep{10.1145/3459637.3481924}. 
Therefore, besides automatic POI verification, it is also important to explore automatic methods for effectively adding, deleting, and updating POIs.
As such, we could provide users with up-to-date POI information to help them make informed decisions about local businesses.

In the future, we plan to extend the capability of \DuMP to maintain the correctness of the large-scale POI database at Baidu Maps from three perspectives, as follows.

\begin{itemize}
    \item \textit{Automatic POI addition}: If a real-world counterpart cannot be found in the database, it is most likely a newly-established POI. Furthermore, to add the POI into the database, \DuMP should recognize the name from its signboard image and calculate the exact coordinates of this POI with the street-view imagery. 
    
    \item \textit{Automatic POI deletion}: Once the coordinates of a newly-added POI are occupied by a certain archived POI in the database, it is very likely that the archived POI has been closed down.  \DuMP should be able to delete this archived POI from the database by analyzing the street views submitted by crowdworkers. 

    \item \textit{Automatic POI updates}: The attributes of a POI, such as its business hours or contact details, may be changed over time. These kinds of information may also appear in the street views in the form of sticker ads. Therefore, we need to extend the capability of \DuMP in extracting more attributes of real-world counterparts from street views \citep{duarus2022,duare2022,dutraffic2022}.
\end{itemize}

\clearpage
\balance 
\bibliographystyle{ACM-Reference-Format}
\bibliography{main}

\end{document}